\documentclass[runningheads]{llncs}

\usepackage{graphicx}
\usepackage{hyperref}
\usepackage{amsmath}
\usepackage{amsfonts}
\usepackage{booktabs}
\usepackage{upgreek}

\begin{document}

\title{SynthA1c: Towards Clinically Interpretable Patient Representations for Diabetes Risk Stratification}

\titlerunning{SynthA1c: Patient Representations for Diabetes Risk Stratification}

\author{Michael S. Yao\inst{\dagger,1,2}\orcidID{0000-0002-7008-6028} \and
Allison Chae\inst{\dagger,2}\orcidID{0000-0001-7029-556X} \and
Matthew T. MacLean\inst{3}\orcidID{0000-0002-0514-7218} \and
Anurag Verma\inst{4}\orcidID{0000-0002-5063-9107} \and
Jeffrey Duda\inst{3}\orcidID{0000-0002-5031-5735} \and
James C. Gee\inst{3}\orcidID{0000-0002-2258-0187} \and
Drew A. Torigian\inst{3}\orcidID{0000-0001-8999-9735} \and
Daniel Rader\inst{4}\orcidID{0000-0002-9245-9876} \and
Charles E. Kahn Jr.\inst{2, 3}\orcidID{0000-0002-6654-7434} \and
Walter R. Witschey\inst{\ddagger,2,3}\orcidID{0000-0003-1669-2120} \and
Hersh Sagreiya\inst{\ddagger,2,3,\thanks{Corresponding author: \texttt{hersh.sagreiya@pennmedicine.upenn.edu}\newline\inst{\dagger} Denotes equal contribution. \hspace{6ex}\inst{\ddagger} Denotes equal contribution.}}\orcidID{0000-0002-2909-6793}}

\authorrunning{M.S. Yao and A. Chae et al.}

\institute{Department of Bioengineering, University of Pennsylvania, Philadelphia PA 19104, USA \and
Perelman School of Medicine, University of Pennsylvania, Philadelphia PA 19104, USA \and
Department of Radiology, University of Pennsylvania, Philadelphia PA 19104 \and
Department of Medicine, University of Pennsylvania, Philadelphia PA 19104}

\maketitle

\begin{abstract}
Early diagnosis of Type 2 Diabetes Mellitus (T2DM) is crucial to enable timely therapeutic interventions and lifestyle modifications. As the time available for clinical office visits shortens and medical imaging data become more widely available, patient image data could be used to opportunistically identify patients for additional T2DM diagnostic workup by physicians. We investigated whether image-derived phenotypic data could be leveraged in tabular learning classifier models to predict T2DM risk in an automated fashion to flag high-risk patients \textit{without} the need for additional blood laboratory measurements. In contrast to traditional binary classifiers, we leverage neural networks and decision tree models to represent patient data as `SynthA1c' latent variables, which mimic blood hemoglobin A1c empirical lab measurements, that achieve sensitivities as high as 87.6\%. To evaluate how SynthA1c models may generalize to other patient populations, we introduce a novel generalizable metric that uses vanilla data augmentation techniques to predict model performance on input out-of-domain covariates. We show that image-derived phenotypes and physical examination data together can accurately predict diabetes risk as a means of opportunistic risk stratification enabled by artificial intelligence and medical imaging. Our code is available at \href{https://github.com/allisonjchae/DMT2RiskAssessment}{https://github.com/allisonjchae/DMT2RiskAssessment}.

\end{abstract}

\section{Introduction}
Type 2 Diabetes Mellitus (T2DM) affects over 30 million patients in the United States, and is most commonly characterized by elevated serum hemoglobin A1c (HbA1c) levels measured through a blood sample \cite{ref1,ref2}. Formally, a patient is considered diabetic if their HbA1c is greater than 6.5\% A1c. While patients diagnosed with T2DM are at an increased risk of many comorbidities, early diagnosis and lifestyle interventions can improve patient outcomes~\cite{ref3}.

However, delayed diagnosis of T2DM is frequent due to a low rate of screening. Up to a third of patients are not screened for T2DM as recommended by current national guidelines \cite{ref4,ref5}, and Porter et al.~\cite{ref6} estimate that it would take over 24 hours per day for primary care physicians to follow national screening recommendations for every adult visit. Furthermore, T2DM screening using patient bloodwork is not routinely performed in acute urgent care settings or emergency department (ED) visits. Given these obstacles, machine learning (ML) is a promising tool to predict patient risk of T2DM and other diseases~\cite{ref7}.

Simultaneously, the usage of radiologic imaging in clinical medicine continues to increase every year \cite{ref8,ref9}. Over 70 million computed tomography (CT) scans are performed annually and their utilization has become increasingly common in both primary care and ED visits~\cite{ref10}. Consequently, the wealth of CT radiographic data can potentially be used to estimate patient risk of T2DM as an incidental finding in these clinical settings. For example, T2DM risk factors include central adiposity and the buildup of excess fat in the liver that can be readily estimated from clinical CT scans. Liver fat excess can be estimated by calculating the spleen-hepatic attenuation difference (SHAD), which is the difference between liver and spleen CT attenuation~\cite{ref11}. These metrics are examples of \textbf{image-derived phenotypes} (IDPs) derived from patient CT scans and other imaging modalities. Other IDPs, such as volume estimation of subcutaneous fat and visceral fat, can also be used to quantify central adiposity. Using these metrics, a prediction model could report estimated T2DM risk as an incidental finding during an unrelated outpatient imaging study or ED visit workup as a means of opportunistic risk stratification from analysis of CT scans and patient information, with automated referral of high-risk patients for downstream screening without the need for intermediate physician intervention.

Existing machine learning methods for disease prediction have largely focused on developing classification models that output probability values for different physiologic states \cite{ref12,ref13,ref14}. However, these metrics are difficult for clinicians to interpret at face value and cannot be intelligently integrated into existing clinician workflows, such as diagnostic pathways based on clinical lab findings~\cite{ref16}.

\begin{figure}
\centering
\includegraphics[width=\textwidth]{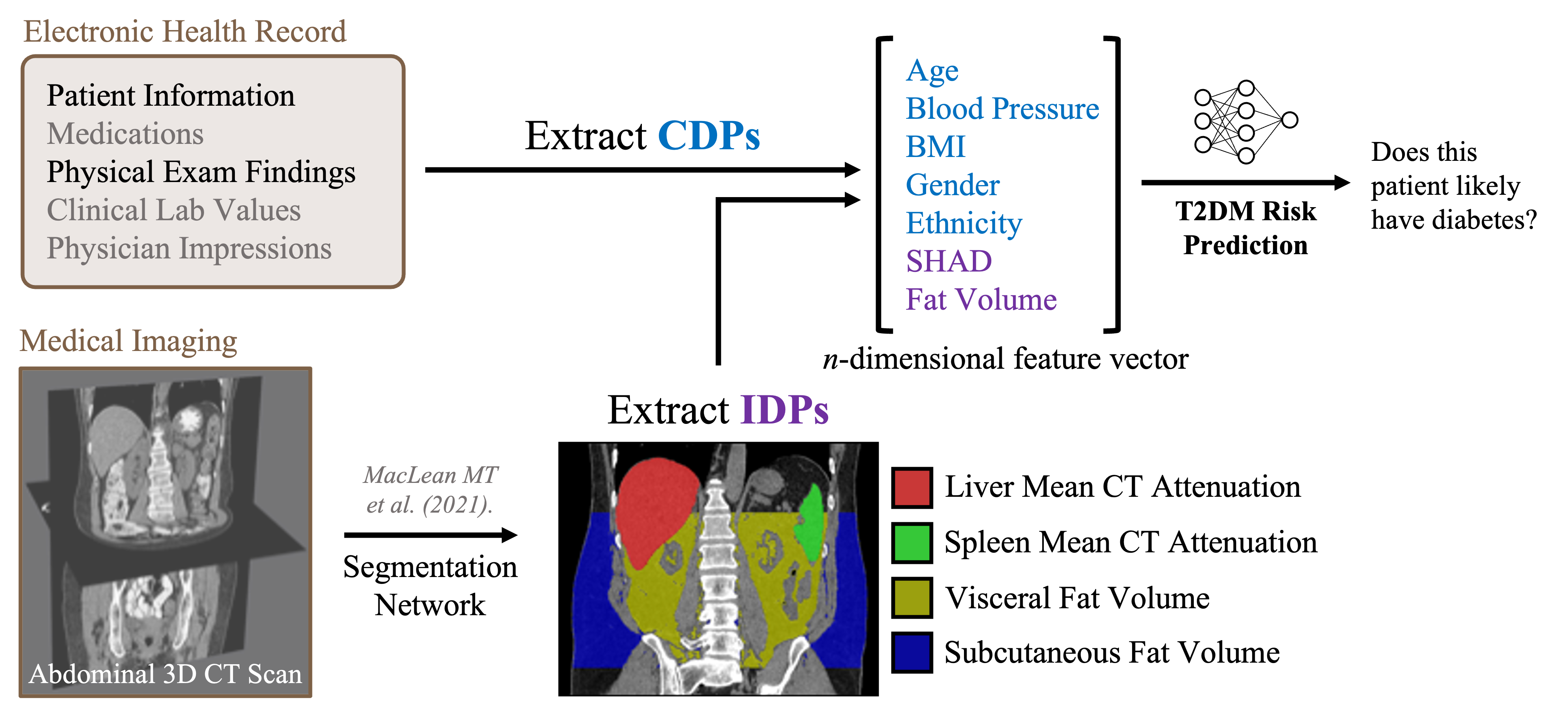}
\caption{Overview of our proposed work. Using the IDP extraction pipeline from MacLean et al.~\cite{ref11}, we can estimate quantitative IDPs from abdominal CT scans associated with an increased risk for T2DM. IDPs and CDPs from corresponding patient electronic health records can then be used to train T2DM risk prediction models. \textit{IDP}: image-derived phenotype; \textit{CT}: computed tomography; \textit{T2DM}: Type 2 Diabetes Mellitus; \textit{CDP}: clinically derived phenotype.} \label{fig1}
\end{figure}

In this study, we hypothesized that radiomic metrics derived from CT scans could be used in conjunction with physical examination data to predict patient T2DM risk using SynthA1c, a novel synthetic \textit{in silico} measurement approximating patient blood hemoglobin A1c (HbA1c) (Fig. \ref{fig1}). To predict model generalizability, we also propose a generalizable data augmentation-based model smoothness metric that predicts SynthA1c accuracy on previously unseen out-of-domain patient datasets.

\section{Materials and Methods}

\subsection{Patient Cohort and Data Declaration}
The data used for our retrospective study were made available by the Penn Medicine BioBank (PMBB), an academic biobank established by the University of Pennsylvania. All patients provided informed consent to utilization of de-identified patient data, which was approved by the Institutional Review Board of the University of Pennsylvania (IRB protocol 813913). From the PMBB outpatient dataset, we obtained patient ages, genders, ethnicities, heights, weights, blood pressures, abdominal CT scans, and blood HbA1c measurements. Notably, the only laboratory value used was HbA1c as a ground truth metric in model training and evaluation\textemdash no blood biomarkers were used as model inputs.

From the PMBB outpatient dataset, we obtained patient ages, genders, ethnicities, heights, weights, blood pressures, abdominal CT scans, and blood HbA1c measurements. Notably, the only clinical laboratory value extracted was HbA1c to be used as a ground truth\textemdash no blood biomarkers were used as model inputs. Patients with any missing features were excluded.

Using the pre-trained abdominal CT segmentation network trained and reported by MacLean et al.~\cite{ref11}, we estimated four IDPs from any given CT of the abdomen and pelvis study (either with or without contrast) to be used as model inputs. Our four IDPs of interest were mean liver CT attenuation, mean spleen CT attenuation, and estimated volume of subcutaneous fat and visceral fat. Briefly, their segmentation network achieved mean S\o renson-Dice coefficients of at least 98\% for all IDP extraction tasks assessed (including our four IDPs of interest) and is detailed further in their work.

Any patient $i$ has a set of measured values of any particular feature within the dataset. To construct a feature vector $\mathbf{x}$ associated with an HbA1c measurement $y_i$, we selected the patient's measurements that minimized the time between the date the feature was measured and the date $y_i$ was measured.

\subsection{Machine Learning Models: GBDT, NODE, and FT-Transformer}
Current supervised methods for disease detection work with feature vectors derived from patient physical examinations and clinical laboratory values \cite{ref7,ref12,ref13}. Our work builds on these prior advances by incorporating IDPs as additional input vector dimensions. Previously, Chen and Guestrin~\cite{ref17} introduced gradient-boosted decision trees (\textbf{GBDTs}) that incorporate scalable gradient boosting with forest classifiers for state-of-the-art prediction accuracy across tasks. A separate class of machine learning models is deep neural networks (DNNs). Recently, neural oblivious decision ensemble (\textbf{NODE}) DNNs achieved classification performance on par with decision tree models on certain tasks~\cite{ref18} and the Feature Tokenizer + Transformer (\textbf{FT-Transformer})~\cite{ref19} effectively adopts transformer architectures to tabular data. Here, we assessed NODE, FT-Transformer, and GBDT architectures as backbones for our SynthA1c encoders.

We sought to compare our proposed SynthA1c models against a number of baselines. We looked at Ordinary Least Squares (OLS) encoders and traditional diabetes \textit{binary classifier} models with the same three architectures as proposed above, in addition to a zero-rule classifier and a multi-rule questionnaire-based classifier currently recommended for clinical practice by the American Diabetes Association and Centers for Disease Control and Prevention~\cite{ref20}.

\subsection{Model Training and Evaluation Strategy}
\label{section23}
Our model inputs can be divided into two disjoint sets: clinically derived phenotypes (CDPs), which are derived from physical examination, and image-derived phenotypes (IDPs) that are estimated from abdominal CT scans herein. The specific CDPs and IDPs used depended on the model class\textemdash broadly, we explored two categories of models, which we refer to as $r$-type and $p$-type. $r$-type models were trained on `raw' data types (CDPs: height, weight, race, gender, age, systolic blood pressure [SBP], diastolic blood pressure [DBP]; IDPs: liver CT attenuation, spleen CT attenuation, subcutaneous fat [SubQ Fat], visceral fat [Visc Fat]), while $p$-type models were trained on `processed' data types (CDPs: BMI, race, gender, age, SBP, DBP; IDPs: SHAD, SubQ Fat, Visc Fat). Comparing the performance of $r$- and $p$- type models could help us better understand if using derivative processed metrics that are better clinically correlated with T2DM risk yields better model performance.

SynthA1c encoders were trained to minimize the $L_2$ distance from the ground truth HbA1c laboratory measurement, and evaluated using the root mean square error (RMSE) and Pearson correlation coefficient (PCC). We then compared the predicted SynthA1c values with the traditional HbA1c $\geq$ 6.5\% A1c diabetes cutoff to assess the utility of SynthA1c outputs in diagnosing T2DM. A $p$ value of $p<0.05$ was used to indicate statistical significance.

\subsection{Implementation Details}
NODE models were trained with a batch size of 16 and a learning rate of $\eta=0.03$, which decayed by half every 40 epochs for a total of 100 epochs. FT-Transformer models were trained with a batch size of 128 and a learning rate of $\eta=0.001$, which decayed by half every 50 epochs for a total of 100 epochs. GBDT models were trained using 32 boosted trees with a maximum tree depth of 8 with a learning rate of $\eta=0.1$.

\subsection{Assessing Out-of-Domain Performance}
An important consideration in high-stakes clinical applications of machine learning is the generalizability of T2DM classifiers to members of previously unseen patient groups. Generalizability is traditionally difficult to quantify and can be affected by training data heterogeneity and the geographic, environmental, and socioeconomic variables unique to the PMBB dataset.

Prior work has shown that model smoothness can be used to predict out-of-domain generalization of neural networks \cite{ref21,ref22}. However, these works largely limit their analysis to classifier networks. To evaluate SynthA1c encoder robustness, we develop an estimation of model manifold smoothness $\mathbb{M}$ for our encoder models. Under the mild assumption that our SynthA1c encoder function $y: \mathbb{R}^{|\mathbf{x}|}\rightarrow \mathbb{R}$ is Lipschitz continuous, we can define a local manifold smoothness metric $\mu$ at $\mathbf{x}=\tilde{\mathbf{x}}$ given by
\begin{equation}
\begin{split}
\mu(\tilde{\mathbf{x}})&=\mathbb{E}_{\mathcal{N}(\tilde{\mathbf{x}})}\left[\frac{\sigma_y^{-1}||y(\mathbf{x})-y(\tilde{\mathbf{x}})||_1}{||\delta \mathbf{x}\oslash \sigma_{\mathbf{x}}||_2}\right]\\
&=\mathcal{V}[\mathcal{N}(\tilde{\mathbf{x}})]^{-1} \cdot \oint_{\mathcal{N}(\tilde{\mathbf{x}})\in\mathcal{D}} d\mathbf{x}\text{ }\frac{\sigma_{y}^{-1}|y(\mathbf{x})-y(\tilde{\mathbf{x}})|}{\sqrt{(\delta \mathbf{x}\oslash \sigma_{\mathbf{x}})^T(\delta \mathbf{x}\oslash \sigma_{\mathbf{x}})}}
\end{split}
\end{equation}
where we have a feature vector $\mathbf{x}$ in domain $\mathcal{D}$ and a neighborhood $\mathcal{N}(\tilde{\mathbf{x}})\in\mathcal{D}$ around $\mathbf{x}$ with an associated volume of $\mathcal{V}[\mathcal{N}(\tilde{\mathbf{x}})]$. We also define $\delta \mathbf{x}=\mathbf{x}-\tilde{\mathbf{x}}$, $\oslash$ as the Hadamard division operator, and $\sigma_{\mathbf{x}}$ as the vector of the estimated standard deviations of each feature over $\mathcal{D}$. The exact expectation value over a given neighborhood $\mathcal{N}(\tilde{\mathbf{x}})$ is computationally intractable, but we can approximate it with a Monte Carlo integration through an empirical sampling of $Q\gg 1$ random feature points $\mathbf{x}_k$ from $\mathcal{N}(\tilde{\mathbf{x}})$:
\begin{equation}
\begin{split}
\mu(\tilde{\mathbf{x}})=\frac{1}{Q}\sum_{k=1}^{Q} \frac{\sigma_{y}^{-1}|y(\mathbf{x}_k)-y(\tilde{\mathbf{x}})|}{\sqrt{(\delta \mathbf{x}_k\oslash \sigma_{\mathbf{x}})^T(\delta \mathbf{x}_k\oslash \sigma_{\mathbf{x}})}}
\end{split}
\end{equation}
We can now define a metric $\mathbb{M}$ for the global encoder manifold smoothness over a domain $\mathcal{D}$ as the expectation value of $\mu(\tilde{\mathbf{x}})$ over $\mathcal{D}$, which can similarly be approximated by an empirical sampling of $N$ feature vectors $\mathbf{x}_1, \mathbf{x}_2, \ldots, \mathbf{x}_N\in\mathcal{D}$. We hypothesized that this global smoothness metric $\mathbb{M}$ inversely correlates with model performance on out-of-domain datasets. To evaluate this experimentally, we assessed model performance on two previously unseen T2DM datasets: (1) the Iraqi Medical City Hospital dataset~\cite{ref23}; and (2) the PMBB inpatient dataset. The Iraqi dataset contains 1,000 sets of patient age, gender, BMI, and HbA1c measurements. Because of this limited feature set, we trained additional SynthA1c encoders (referred to as $p$'-type models) on the PMBB outpatient dataset using only these features. The PMBB inpatient dataset consists of 2,066 measurements of the same datatypes as the outpatient dataset (Section \ref{section31}).

\section{Results}

\subsection{Summary Statistics}
\label{section31}
Our model-building dataset from the PMBB consisted of 2,077 unique HbA1c measurements (1,159 diabetic, 619 prediabetic, 299 nondiabetic) derived from 389 patients (Table \ref{tab1}). 208 (10\%) samples were set aside as a holdout test set partition disjoint by patient identity. Each HbA1c measurement was used to construct an associated feature vector from that patient's data collected closest in time to each HbA1c measurement. To quantify the temporal association between a given patient's measurements, we defined the daterange of an observation vector $\mathbf{x}$ as the maximum length of time between any two features/imaging studies. The median daterange in our dataset was 18 days.

\begin{table}
  % \caption{Population characteristics for PMBB outpatient cohort. Values are $n$ (\%).}\label{tab1}
  \caption{PMBB outpatient dataset characteristics. To reduce the effects of selection bias, all patients presenting to the University of Pennsylvania Health System were given the opportunity to enroll in the PMBB so as to best capture the population of patients that seek medical care and avoid overrepresentation of healthy patients as in traditional office visit patient recruitment strategies. However, the PMBB is still affected by hesitancies of patient sub-populations in study enrollment and the unique socioeconomic factors affecting different groups of patients. \textit{HTN}: Hypertension.}\label{tab1}
  \centering
  \resizebox{\textwidth}{!}{\begin{tabular}{lc}
    \toprule
    \hspace{2ex}\textbf{Self-Reported Ethnicity} & \textbf{Count (\%)}\\
    % \midrule
      \hspace{4ex}White\hspace{20ex} & 720 (34.7) \\
      \hspace{4ex}Hispanic\hspace{20ex} & 40 (1.9) \\
      \hspace{4ex}Black\hspace{20ex} & 1248 (60.1) \\
      \hspace{4ex}Asian\hspace{20ex} & 36 (1.7) \\
      \hspace{4ex}Pacific Islander\hspace{20ex} & 6 (0.3) \\
      \hspace{4ex}Native American\hspace{20ex} & 5 (0.2) \\
      \hspace{4ex}Other/Unknown\hspace{20ex} & 22 (1.1) \\
    % \bottomrule
    % \toprule
    \hspace{2ex}\textbf{Self-Reported Gender} & \textbf{Count (\%)}\\
    % \midrule
      \hspace{4ex}Male\hspace{20ex} & 880 (42.4) \\
      \hspace{4ex}Female\hspace{20ex} & 1197 (57.6) \\
    % \bottomrule
    % \toprule
    \hspace{2ex}\textbf{Age Decade} & \textbf{Count (\%)}\\
    % \midrule
      \hspace{4ex}20-29\hspace{20ex} & 31 (1.5) \\
      \hspace{4ex}30-39\hspace{20ex} & 89 (4.3) \\
      \hspace{4ex}40-49\hspace{20ex} & 362 (17.4) \\
      \hspace{4ex}50-59\hspace{20ex} & 593 (28.6) \\
      \hspace{4ex}60-69\hspace{20ex} & 680 (32.7) \\
      \hspace{4ex}70-79\hspace{20ex} & 299 (14.4) \\
      \hspace{4ex}80-89\hspace{20ex} & 23 (1.1) \\
    % \bottomrule
    % \toprule
    \hspace{2ex}\textbf{Blood Pressure} & \textbf{Count (\%)}\\
    % \midrule
      \hspace{4ex}Normal (SBP $<$ 120 mmHg and DBP $<$ 80 mmHg) & 421 (20.2)\\
      \hspace{4ex}Elevated (120 $\leq$ SBP $<$ 130 mmHg and DBP $<$ 80 mmHg) & 398 (19.2)\\
      \hspace{4ex}Stage 1 HTN (130 $\leq$ SBP $<$ 140 mmHg or 80 $\leq$ DBP $<$ 90 mmHg) & 652 (31.4)\\
      \hspace{4ex}Stage 2 HTN (SBP $\geq$ 140 mmHg or DBP $\geq$ 90 mmHg) & 606 (29.2)\\
    % \bottomrule
    % \toprule
    \hspace{2ex}\textbf{BMI} & \textbf{Count (\%)}\\
    % \midrule
      \hspace{4ex}Underweight or Healthy Weight (BMI $<$ 25.0) & 275 (13.2)\\
      \hspace{4ex}Overweight (25.0 $\leq$ BMI $<$ 30.0) & 443 (21.3)\\
      \hspace{4ex}Class 1 Obesity (30.0 $\leq$ BMI $<$ 35.0) & 556 (26.8)\\
      \hspace{4ex}Class 2 Obesity (35.0 $\leq$ BMI $<$ 40.0) & 389 (18.7)\\
      \hspace{4ex}Class 3 Obesity (BMI $\geq$ 40.0) & 414 (20.0)\\
    % \bottomrule
    % \toprule
    \hspace{2ex}\textbf{HbA1c} & \textbf{Count (\%)}\\
    % \midrule
      \hspace{4ex}Not Diabetic (HbA1c $<$ 6.5\% A1c) & 918 (44.2)\\
      \hspace{4ex}Diabetic (HbA1c $\geq$ 6.5\% A1c) & 1159 (55.8)\\
    % \bottomrule
    % \toprule
    \hspace{2ex}\textbf{CT Abdomen and Pelvis Enhancement} & \textbf{Count (\%)}\\
    % \midrule
      \hspace{4ex}With Contrast & 1570 (75.6)\\
      \hspace{4ex}Without Contrast & 507 (24.4)\\
    % \bottomrule
    % \toprule
    \hspace{2ex}\textbf{Image Derived Phenotypes (IDPs) Statistics} & \textbf{Mean $\pm$ SD}\\
    % \midrule
      \hspace{4ex}Spleen CT Attenuation (HU) & 36.2 $\pm$ 16.7\\
      \hspace{4ex}Liver CT Attenuation (HU) & 42.8 $\pm$ 20.2\\ 
      \hspace{4ex}Subcutaneous Fat Area (cm\textsuperscript{2}) & 321.3 $\pm$ 170.1\\
      \hspace{4ex}Visceral Fat Area (cm\textsuperscript{2}) & 172.4 $\pm$ 104.9\\
    % \bottomrule
    \toprule
    \hspace{2ex}\textbf{Total Count} & \textbf{2077}\\
    \bottomrule
  \end{tabular}}
\end{table}

\subsection{SynthA1c Encoder Experimental Results}
Our results suggest that the GBDT encoder predicted SynthA1c values closest to ground truth HbA1c values, followed by both the NODE and FT-Transformer DNN models (Table \ref{tab2}). All the learning-based architectures assessed outperformed the baseline OLS encoder. When comparing SynthA1c outputs against the clinical HbA1c cutoff of 6.5\% A1c for the diagnosis of diabetes, the $r$-GBDT SynthA1c model demonstrated the highest sensitivity of the assessed models at 87.6\% on par with the best-performing binary classifier model assessed. In terms of an opportunistic screening modality for T2DM, a high sensitivity ensures that a large proportion of patients with diabetes can be identified for additional lab-based diagnostic work-up with their primary care physicians. Although the accuracy of SynthA1c encoders was lower than the corresponding binary classifier models assessed, this may be partially explained by the fact that the latter's threshold value for classification was empirically tuned to maximize the model's accuracy. In contrast, our SynthA1c encoders used the fixed clinical HbA1c cutoff of 6.5\% A1c for diabetes classification. When comparing $r$- and $p$- type SynthA1c models, we did not observe a consistently superior data representation strategy.

\begin{table}[ht]
  \caption{SynthA1c prediction results using different encoder models. $r$- ($p$-) prefixed models are fed raw (processed) inputs as outlined in Section \ref{section23}. RMSE in units of \% A1c. For the SynthA1c encoder models, recall, precision, specificity, and accuracy metrics are reported based on the traditional T2DM cutoff of 6.5\% A1c. The Multi-Rule binary classifier is the current deterministic risk stratification tool recommended by American Diabetes Association~\cite{ref20}.}\label{tab2}
  \centering
  \begin{tabular}{rcccccc}
    \toprule
    \textbf{SynthA1c Encoder} & RMSE & PCC & Recall & Precision & Specificity & Accuracy\\
    \midrule
    $r$-OLS & 1.67 & 0.206 & 85.3 & 56.0 & 26.3 & 57.2\\
    $p$-OLS & 1.73 & 0.159 & 80.7 & 57.5 & 34.3 & 58.6\\
    $r$-FT-Transformer & 1.44 & 0.517 & 87.6 & 63.4 & 55.9 & 70.7\\
    $p$-FT-Transformer & 1.51 & 0.441 & 83.5 & 61.4 & 54.1 & 67.8\\
    $r$-NODE & 1.60 & 0.378 & 85.6 & 55.0 & 38.7 & 60.6\\
    $p$-NODE & 1.57 & 0.649 & 77.3 & 59.5 & 54.1 & 64.9\\
    $r$-GBDT & 1.36 & 0.567 & 87.2 & 66.4 & 51.5 & 70.2\\
    $p$-GBDT & 1.36 & 0.591 & 77.1 & 72.4 & 67.7 & 72.6\\
    \bottomrule
    \toprule
    \textbf{Binary Classifier} & \multicolumn{2}{c}{AUROC (\%)} & Recall & Precision & Specificity & Accuracy\\
    \midrule
    Zero-Rule & \multicolumn{2}{c}{\textemdash} & 100 & 52.4 & 0.0 & 52.4\\
    Multi-Rule & \multicolumn{2}{c}{56.3} & 67.0 & 54.9 & 39.4 & 53.8\\
    $r$-FT-Transformer & \multicolumn{2}{c}{82.1} & 85.3 & 73.8 & 66.7 & 76.4\\
    $r$-NODE & \multicolumn{2}{c}{83.5} & 82.6 & 76.9 & 72.7 & 77.9\\
    $r$-GBDT & \multicolumn{2}{c}{83.1} & 87.2 & 76.6 & 70.7 & 79.3\\
    \bottomrule
  \end{tabular}
\end{table}

\begin{figure}
\includegraphics[width=\textwidth]{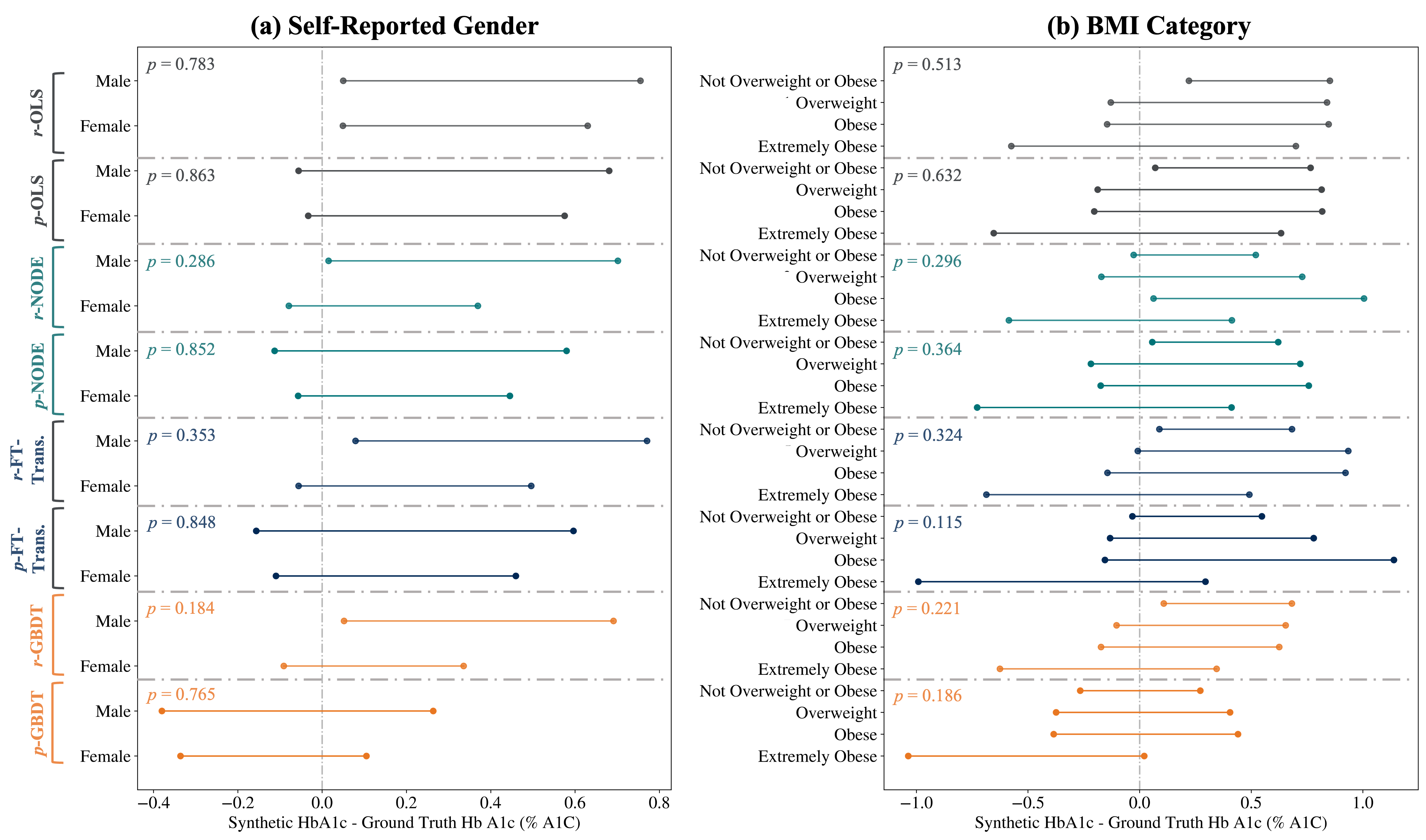}
\caption{Assessing for algorithmic bias in SynthA1c encoders. We plotted the 95\% confidence interval of the mean difference between the SynthA1c model output and ground truth HbA1c as a function of self-reported (a) gender and (b) BMI category. $p$ values comparing the differences in SynthA1c model performance when stratified by gender (two-sample T-test) and BMI category (one-way ANOVA) are shown.} \label{fig2}
\end{figure}

To further interrogate our SynthA1c encoders, we investigated whether model performance varied as a function of demographic features. Defining the difference between the model prediction and ground truth HbA1c values as a proxy for model performance, all SynthA1c encoders showed no statistically significant difference in performance when stratified by gender or BMI (Fig. \ref{fig2}).

\subsection{Ablation Studies: Relative Importance of CDPs and IDPs}
Until now, prior T2DM classifiers have used only blood lab measurements and physical examination data to predict T2DM. In contrast, our models presented herein are the first to incorporate IDPs as input model features for the task of diabetes risk stratification. To better understand the benefit and value-add of using IDPs in conjunction with CDPs, we evaluated classifier performance on models trained using either only CDPs or only IDPs and compared them to corresponding models trained using both input types.

Our results suggest that while classifier models trained only on CDPs generally outperform those trained only on IDPs, the best performance is achieved when combining CDPs and IDPs together (Table \ref{tab3}). This further validates the clinical utility of incorporating IDPs into patient diagnosis and disease risk stratification first proposed by MacLean et al.~\cite{ref11} and related work.

\begin{table}[ht]
  \caption{Ablation study assessing model performance as a function of clinically derived phenotypes (CDPs) and/or image-derived phenotypes (IDPs).}\label{tab3}
  \centering
  \begin{tabular}{rcccc}
    \toprule
    \textbf{$r$-NODE} & Recall & Precision & Specificity & Accuracy\\
    \midrule
    CDPs Only & 77.1 & 73.7 & 69.7 & 73.5\\
    IDPs Only & 73.4 & 76.9 & 75.8 & 74.5\\
    CDPs + IDPs & 82.6 & 76.9 & 72.7 & 77.9\\
    \bottomrule
    \toprule
    \textbf{$r$-FT-Transformer} & Recall & Precision & Specificity & Accuracy\\
    \midrule
    CDPs Only & 78.0 & 76.6 & 73.7 & 75.9\\
    IDPs Only & 71.6 & 60.5 & 48.5 & 60.6\\
    CDPs + IDPs & 85.3 & 73.8 & 66.7 & 76.4\\
    \bottomrule
    \toprule
    \textbf{$r$-GBDT} & Recall & Precision & Specificity & Accuracy\\
    \midrule
    CDPs Only & 80.7 & 68.6 & 59.6 & 70.7\\
    IDPs Only & 73.4 & 75.5 & 73.7 & 73.6\\
    CDPs + IDPs & 87.2 & 76.6 & 70.7 & 79.3\\
    \bottomrule
  \end{tabular}
\end{table}

\subsection{Characterizing Out-of-Domain Model Performance}
As our proposed global smoothness metric $\mathbb{M}$ decrease across the three evaluated models, the RMSE in SynthA1c prediction decreases and the PCC increases, corresponding to better predictive performance on the out-of-domain Iraqi Medical Center Dataset (Table \ref{tab4}). This supports our initial hypothesis that smoother models may generalize better to unseen datasets. We also noted larger RMSE values using the Iraqi Medical Center Dataset when compared to the PMBB outpatient test dataset results (Table \ref{tab2}).

\begin{table}[ht]
  \caption{SynthA1c model sensitivity and out-of-domain generalization results. Global smoothness metric values $\mathbb{M}$ were evaluated on the PMBB outpatient dataset. $r$-type models could not be evaluated on the Iraqi dataset because IDPs and medical imaging data were not available. RMSE in units of \% A1c.}\label{tab4}
  \centering
  \begin{tabular}{rcccccc}
    \toprule
     &  & \multicolumn{2}{c}{Iraqi Dataset} & & \multicolumn{2}{c}{PMBB Inpatient}\\
    \cmidrule{3-4} \cmidrule{6-7}
    SynthA1c Encoder & $\mathbb{M}$ & RMSE & PCC & & RMSE & PCC\\
    \midrule
    $p$'-/$r$- NODE & 1.43 & 3.62 / \textemdash & 0.154 / \textemdash & & 1.76 / 1.23 & 0.512 / 0.795\\
    $p$'-/$r$- FT-Transformer & 1.07 & 3.04 / \textemdash & 0.246 / \textemdash & & 1.90 / 1.58 & 0.331 / 0.617\\
    $p$'-/$r$- GBDT & 3.28 & 6.25 / \textemdash & 0.021 / \textemdash & & 1.54 / 1.12 & 0.674 / 0.823\\
    \bottomrule
  \end{tabular}
\end{table}

Interestingly, we found that this relationship did not ostensibly hold when considering the PMBB inpatient dataset; in fact, model predictive performance was \textit{inversely} correlated with global smoothness. This suggested that the PMBB inpatient and outpatient dataset distributions were more similar than initially predicted. To validate this hypothesis, we computed the Kullback-Leibler (KL) divergence between each of the test dataset distributions and the training dataset distribution with respect to the features available in all datasets: ethnicity, gender, age, BMI, and HbA1c. We assumed the PMBB-derived outpatient training dataset was sampled from a distribution $\mathcal{Q}$ and each of the PMBB outpatient test, PMBB inpatient, and Iraqi medical center datasets were sampled from $\mathcal{P}_{\text{Outpatient}}$, $\mathcal{P}_{\text{Inpatient}}$, and $\mathcal{P}_{\text{Iraqi}}$, respectively. The greatest KL divergence was between the Iraqi medical center and training dataset distributions, as expected ($D_{KL}[\mathcal{P}_{\text{Iraqi}} || \mathcal{Q}] = 31.2$). Despite the fact that our training set included outpatient data alone, we found the KL divergence between the inpatient test and training datasets ($D_{KL}[\mathcal{P}_{\text{Inpatient}} || \mathcal{Q}] = 0.227$) was lower than that between the outpatient test and training dataset ($D_{KL}[\mathcal{P}_{\text{Outpatient}} || \mathcal{Q}] = 1.84$).

To further characterize the feature distributions within our datasets, we analyzed the pairwise relationships between BMI, age, and HbA1c. Individual feature distributions were statistically significant between either of the PMBB datasets and the Iraqi Medical Center dataset (two-sample Kolmogorov-Smirnov (KS) test; $p < 0.0001$ between [PMBB inpatient dataset, Iraqi Medical Center dataset] and [PMBB outpatient dataset, Iraqi Medical Center dataset] pairs for individual age, HbA1c, and BMI quantitative features), but not between the PMBB inpatient and outpatient datasets (two-sample KS test; age: $p = 0.315$, HbA1c: $p = 0.463$, BMI: $p = 0.345$). These results suggest that inpatients are a compact subset of outpatients within the PMBB with respect to T2DM risk assessment. This helps explain our initial findings regarding the relationship between $\mathbb{M}$ and model generalization. Further work is warranted to validate the proposed metric $\mathbb{M}$ across other tasks.

\section{Conclusion}
Our work highlights the value of using CT-derived IDPs and CDPs for opportunistic screening of T2DM. We show that tabular learning architectures can act as novel SynthA1c encoders to predict HbA1c measurements noninvasively. Furthermore, we demonstrate that model manifold smoothness may be correlated with prediction performance on previously unseen data sampled from out-of-domain patient populations, although additional validation studies on separate tasks are needed. Ultimately, we hope that our proposed work may be used in every outpatient and ED imaging study, regardless of chief complaint, for opportunistic screening of type 2 diabetes. Our proposed SynthA1c methodology will by no means replace existing diagnostic laboratory workups, but rather identify those at-risk patients who should consider consulting their physician for downstream clinical evaluation in an efficient and automated manner.

% Future work remains to be done to improve the predictive power of learning-based patient T2DM status prediction. Firstly, the increasingly widespread availability of clinical imaging data enables potential work in tracking changes in patient-specific imaging studies and extracted IDPs over periods of time, which could be indicative of disease progression or remission. Such time series data could represent powerful additional features that better take into the account the longitudinal effects of therapeutic regimens, lifestyle modifications, and other factors. For example, prior work by Pimentel et al.~\cite{ref24} showed that time series analysis using physical exam features alone could achieve comparable predictive power compared to corresponding models trained on static feature vectors consisting of both physical exam and more invasive clinical lab measurements.

% Medical biobanks, such as the Penn Medicine BioBank used to construct our dataset herein, also feature other information-rich data types in addition to imaging and clinical data. Modern health datasets now include genomic sequencing, wearable devices, and histological slides among many others that can enable powerful advancements in precision medicine in addition to disease and prognosis forecasting. As such data become increasingly available, multimodal ML methods that are able to effectively incorporate all these data types may improve upon current techniques.

\subsubsection{Acknowledgments}
MSY is supported by NIH T32 EB009384. AC is supported by the A$\Upomega$A Carolyn L. Kuckein Student Research Fellowship and the University of Pennsylvania Diagnostic Radiology Research Fellowship. WRW is supported by NIH R01 HL137984. MTM received funding from the Sarnoff Cardiovascular Research Foundation. HS received funding from the RSNA Scholar Grant.

% \section*{Author Contributions and Competing Interests}
% MTM, DR, and WRW, obtained funding for the study. MSY, AC, WRW, and HS were responsible for the concept and design of the study. WRW and HS supervised and coordinated the research. MTM, DR, and WRW were involved in patient identification and data procurement from the clinical workflow. MSY and AC performed model training and evaluation, and MSY, AC, WRW, and HS performed the statistical analysis. MSY and AC drafted the manuscript, and all authors revised and approved the final manuscript. The authors declare no competing interests related to this work.


\end{document}